\documentclass{article}

\PassOptionsToPackage{numbers,compress}{natbib}
\usepackage{amsmath}
\usepackage[pdftex]{graphicx} 

\usepackage[final]{neurips_2021}

\usepackage[utf8]{inputenc} % allow utf-8 input
\usepackage[T1]{fontenc}    % use 8-bit T1 fonts
\usepackage{hyperref}       % hyperlinks
\usepackage{url}            % simple URL typesetting
\usepackage{booktabs}       % professional-quality tables
\usepackage{amsfonts}       % blackboard math symbols
\usepackage{nicefrac}       % compact symbols for 1/2, etc.
\usepackage{microtype}      % microtypography
\usepackage{xcolor}         % colors
\usepackage{float}

\title{Reverse engineering recurrent neural networks with Jacobian switching linear dynamical systems}

\author{%
  Jimmy T.H. ~Smith \\
  Institute for Computational and Mathematical Engineering\\
  Stanford University\\
  Stanford, CA 94305 \\
  \texttt{jsmith14@stanford.edu} \\
   \And
   Scott W. Linderman \\
   Department of Statistics \\
   Stanford University\\
   Stanford, CA 94305 \\
   \texttt{scott.linderman@stanford.edu} \\
   \And
   David Sussillo \\
   Department of Electrical Engineering \\
   Stanford University\\
   Stanford, CA 94305 \\
   \texttt{sussillo@stanford.edu} \\
}

\begin{document}

\maketitle

\begin{abstract}
Recurrent neural networks (RNNs) are powerful models for processing time-series data, but it remains challenging to understand how they function. Improving this understanding is of substantial interest to both the machine learning and neuroscience communities. The framework of reverse engineering a trained RNN by linearizing around its fixed points has provided insight, but the approach has significant challenges. These include difficulty choosing which fixed point to expand around when studying RNN dynamics and error accumulation when reconstructing the nonlinear dynamics with the linearized dynamics. We present a new model that overcomes these limitations by co-training an RNN with a novel switching linear dynamical system (SLDS) formulation. A first-order Taylor series expansion of the co-trained RNN and an auxiliary function trained to pick out the RNN's fixed points govern the SLDS dynamics. The results are a trained SLDS variant that closely approximates the RNN, an auxiliary function that can produce a fixed point for each point in state-space, and a trained nonlinear RNN whose dynamics have been regularized such that its first-order terms perform the computation, if possible. This model removes the post-training fixed point optimization and allows us to unambiguously study the learned dynamics of the SLDS at any point in state-space.  It also generalizes SLDS models to continuous manifolds of switching points while sharing parameters across switches. We validate the utility of the model on two synthetic tasks relevant to previous work reverse engineering RNNs. We then show that our model can be used as a drop-in in more complex architectures, such as LFADS, and apply this LFADS hybrid to analyze single-trial spiking activity from the motor system of a non-human primate.
\end{abstract}

\section{Introduction}
Recurrent neural networks (RNNs) are a powerful and popular tool for modeling complex sequence data. They learn to transform input sequences into output sequences by using an internal state that allows data from the past to influence the current state. RNNs have been utilized in various applications such as speech recognition, sentiment analysis, music, and video~\cite{speech_app, video_app,music_app, tang-etal-2015-document}.  In neuroscience, RNNs have been used for modeling large-scale neural recordings and as a generator of scientific hypotheses by studying the network's learned representations~\cite{pandarinath2018inferring,j.2018emergence, NEURIPS2019_6e7d5d25, banino2018vector, Kanitscheider231159,NEURIPS2020_30f0641c}. However, RNNs are generally viewed as black boxes. While there has been progress in understanding their operation on simple tasks, rigorously understanding how they solve complex tasks remains a significant challenge. 

An important line of work to improve our understanding of RNN computations uses dynamical systems theory~\cite{tsung94,6796260,Casey1996TheDO,rodriguez1999recurrent, Sussillo_Barak,Jordan2019GatedRU, fu2019dynamically, nguyen2020variational, li2021fourier}. In particular,~\citet{Sussillo_Barak} proposed reverse engineering a trained RNN by using numerical optimization to find the RNN's fixed and slow points. The RNN is linearized around these points, and the resulting linear approximation dynamics are studied to draw insights into how the RNN solves the task. While there are no guarantees concerning the success of this approach, empirically, linearization around fixed and slow points has led to insights in numerous applications~\cite{NEURIPS2020_30f0641c, mante2013context,NEURIPS2019_d921c3c7,Maheswaranathan2020HowRN, CARNEVALE20151067,sussillo2015neural,NEURIPS2019_5f5d4720,Finkelstein2019.12.14.876425}. There are a few drawbacks of this method. First, it requires a separate numerical optimization routine after training the network. Second, it can be ambiguous which fixed point to linearize around for any given point in state space. The standard fixed point finding numerical optimization routine provides a collection of fixed points but no direct link between these points and locations in state space.  Finally, simulating nonlinear RNNs with linearizations around fixed points can slowly accumulate significant error, forcing previous attempts to resort to one-step ahead dynamics generation~\cite{NEURIPS2019_d921c3c7}. These problems can lead to uncertainty of how well switching between linearizations around fixed and slow points describes the nonlinear dynamics. 

Here, we combine ideas from reverse engineering RNNs and switching linear dynamical systems (SLDS)~\cite{Ackerson_Fu,chang_athans,Murphy98switchingkalman,Ghahramani_Hinton, Fox_slds} to address these challenges.
Given an RNN we would like to train and analyze, we introduce a separate network consisting of a novel SLDS formulation based on the first-order Taylor series expansion of the RNN equation and a learnable auxiliary function that produces the RNN's fixed/slow points. We then define a loss function that includes regularization terms to force the SLDS to approximate the RNN and switch about the RNN's fixed/slow points. After co-training these three functions with standard RNN training methods, the result is an accurate switching linear approximation of the nonlinear RNN and a trained auxiliary function that provides fixed and slow points of the RNN. This architecture and training procedure:
\begin{enumerate} 
\item Eliminates the need for post-training fixed point finding. \item Generalizes SLDS to be able to switch about continuous manifolds of fixed points. 
\item Enables parameter sharing between the SLDS switches. %(potentially discrete)
\item Allows the nonlinear RNN dynamics to be approximated by switching between local linearizations around fixed points, if possible.
\end{enumerate}
The combination of these benefits significantly simplifies the process of reverse engineering an RNN using fixed points. We illustrate the method on two synthetic tasks and a neural dataset. 

\section{Review of reverse engineering RNNs and SLDSs}
\subsection{Reverse engineering RNNs with fixed points}
\label{rev_eng}
The motivation for reverse engineering RNNs with fixed point analysis is the hypothesis that trained RNNs use mechanisms to solve tasks that are described well by the linearized dynamics around its fixed and slow points~\cite{Sussillo_Barak}. These points are state vectors $\mathbf{h}^*\in \mathbb{R}^D$ that, given an input $\mathbf{u}^*\in \mathbb{R}^U$,  do not significantly change when applying the RNN update function,~$\mathbf{F}: \mathbb{R}^D \times \mathbb{R}^U \to \mathbb{R}^D$. That is, $\mathbf{h}^* \approx \mathbf{F}(\mathbf{h}^*, \mathbf{u}^*; \mathbf{\theta})$.  We can find these points numerically by minimizing a loss function
\begin{align}
    \mathcal{L}(\mathbf{h}) = \|\mathbf{h}-\mathbf{F}(\mathbf{h}, \mathbf{u}^*;\mathbf{\theta}) \|_2^2,
\end{align}
using auto-differentiation methods~\cite{Golub2018}. In practice, one typically initializes the candidate fixed points with the hidden states produced from running forward passes given trial inputs. Once we have found the fixed/slow points, we linearize the system around these points and analyze the dynamics of the linearized system to determine how the system computes. This reverse engineering method is supported in theory by the Hartman-Grobman theorem ~\cite{grobman1959homeomorphism, hartman1960lemma, arrowsmith1992dynamical}, which says that the dynamics of a nonlinear system in a domain near a hyperbolic fixed point is qualitatively the same as the dynamics of its linearization near this point. A potential theoretical issue arises with non-hyperbolic fixed points, though empirically, this has not seemed to pose a major issue for this approach.  Finally, we note a recent alternative method to finding fixed points presented in~\cite{katz2017using} that makes use of mathematical objects called directional fibers.

\subsection{Switching linear dynamical systems}
\label{slds}
Models based on a linear dynamical system (LDS) are often used to model multi-dimensional time series and lend themselves well to dynamical systems analyses. The basic LDS models time series data using a latent representation that follows linear dynamics.  A switching LDS (SLDS) augments the basic model with discrete states that correspond to different linear dynamics. This allows a SLDS to break down complex, nonlinear time series into a sequence of simpler local linear dynamics. Explicitly, let $\mathbf{y}_t \in \mathbb{R}^N$ denote the observation data at time $t$ and let $\mathbf{h}_t\in \mathbb{R}^D$ represent the corresponding continuous latent state. A SLDS models the expected observation value as $\mathbb{E}[y_t] = \mathbf{g}(\mathbf{h}_t)$ where $\mathbf{g}$ is a mapping from $\mathbb{R}^D$ to $\mathbb{R}^N$. Given the discrete latent state $z_t\in \{1,...,K\}$ and input $\mathbf{u}_t\in \mathbb{R}^U$, the SLDS continuous latent states follow linear dynamics,
\begin{align}
    \mathbf{h}_t \sim \mathcal{N}\big(A^{(z_t)}\mathbf{h}_{t-1}+V^{(z_t)}\mathbf{u}_t+b^{(z_t)}, Q^{(z_t)} \big). \label{slds_eq}
\end{align}

The current discrete state determines the current linear dynamics $A^{(z_t)}\in \mathbb{R}^{DxD}$, input matrix $V^{(z_t)}\in \mathbb{R}^{DxU}$, bias term $b^{(z_t)}\in \mathbb{R}^D$ and noise covariance $Q^{(z_t)}\in \mathbb{R}^{DxD}$.
In the basic SLDS, a Markov transition matrix generally defines the discrete state switching probability.  An extension to the SLDS model is the recurrent switching linear dynamical system (RSLDS) which allows the discrete state transition to depend on the previous continuous latent state~\cite{pmlr-v54-linderman17a,Linderman621540,nassar2018treestructured, NEURIPS2020_aa1f5f73}. This recurrent connection allows more expressivity in the model and corresponds to the idea that the current discrete state (and therefore the current dynamical regime) should depend on the current state space location.  

SLDS models can often offer a balance between interpretability and expressivity for many problems. For example, one can improve the expressivity by increasing the number of discrete states, but this can come at the cost of interpretability and an increase in learnable parameters. In addition, SLDS generally requires hyperparameter tuning to determine the optimal number of discrete states to use.

\begin{figure}
  \centering
  \includegraphics[width=\linewidth]{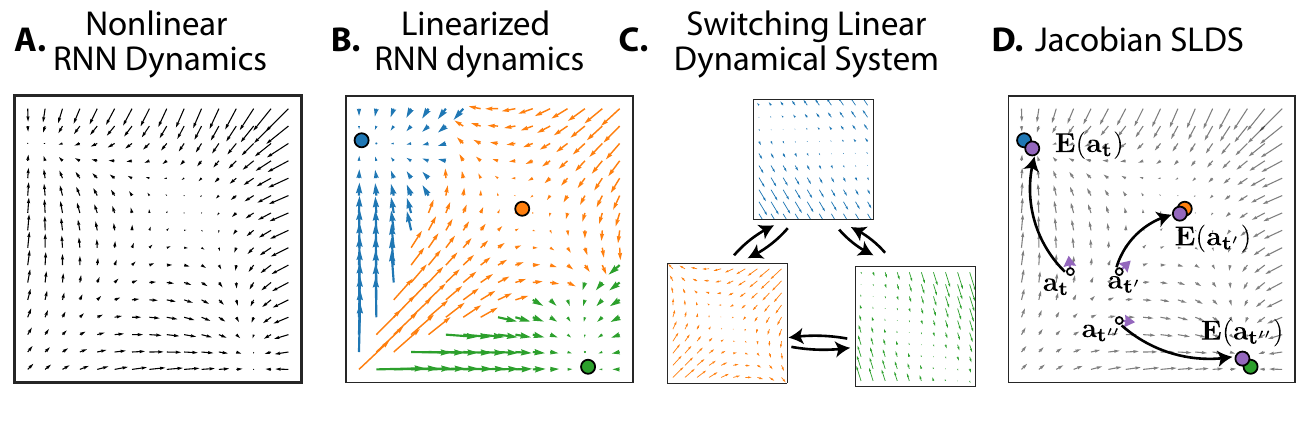}
  \vspace{-1em}
  \caption{\textbf{A}.~Example of a nonlinear dynamical system learned by a recurrent neural network (RNN). \textbf{B}.~Linearization of the RNN dynamics around the nearest fixed point (dots). \textbf{C}.~A switching linear dynamical system (SLDS) approximates the nonlinear dynamics by stochastically jumping between three linear regimes. \textbf{D}.~The Jacobian SLDS co-trains an RNN (gray arrows) with an \textit{expansion network} ($\mathbf{E(\cdot)}$), which maps the current JSLDS state (white dots, $\mathbf{a_t}$) to an expansion point (purple dots, $\mathbf{E(a_t)}$) near to a true fixed point of the RNN. The JSLDS linearizes the RNN dynamics around the expansion point to obtain a linear system (purple arrows) that approximates the local dynamics. }
  \label{graph_model}
\end{figure}

\section{Jacobian Switching Linear Dynamical System}
\label{model}
We now present the Jacobian Switching Linear Dynamical System (JSLDS) model and training procedure.
It combines ideas from reverse engineering RNNs and SLDS to achieve automated fixed point finding and accurate SLDS approximations of nonlinear RNNs.

\subsection{Motivation}
Let $\mathbf{F}$ denote a nonlinear RNN with previous state $\mathbf{h}_{t-1}\in \mathbb{R}^D$, input $\mathbf{u}_t\in \mathbb{R}^U$, and parameters $\mathbf{\theta}$. Writing the RNN update equation and its first order Taylor series expansion around points $\mathbf{h}^*$ and $\mathbf{u}^*$ we have 
\begin{align}
    \mathbf{h}_t &= \mathbf{F}(\mathbf{h}_{t-1}, \mathbf{u}_t; \mathbf{\theta})
    \approx \mathbf{F}(\mathbf{h}^*, \mathbf{u}^*; \mathbf{\theta}) + \frac{\partial\mathbf{F}}{\partial \mathbf{h}}(\mathbf{h}^*, \mathbf{u}^*; \mathbf{\theta})\left(\mathbf{h}_{t-1} - \mathbf{h}^*\right) + \frac{\partial\mathbf{F}}{\partial \mathbf{u}}(\mathbf{h}^*, \mathbf{u}^*; \mathbf{\theta})\left(\mathbf{u}_t-\mathbf{u}^*\right)  \label{fotse}.
\end{align}
Here, $\frac{\partial\mathbf{F}}{\partial \mathbf{h}}(\mathbf{h}^*, \mathbf{u}^*; \mathbf{\theta})\in \mathbb{R}^{DxD}$ is the \textbf{recurrent Jacobian} that determines the recurrent local dynamics and $\frac{\partial\mathbf{F}}{\partial \mathbf{u}}(\mathbf{h}^*, \mathbf{u}^*; \mathbf{\theta})\in \mathbb{R}^{DxU}$ is the \textbf{input Jacobian} that determines the system's input sensitivity.  If $\mathbf{h}^*$ is a fixed point of $\mathbf{F}$ for a given $\mathbf{u}^*$, then $\mathbf{F}(\mathbf{h}^*, \mathbf{u}^*; \mathbf{\theta})=\mathbf{h}^*$ and equation (\ref{fotse}) yields
\begin{align}
        \mathbf{h}_t- \mathbf{h}^*&\approx \frac{\partial\mathbf{F}}{\partial \mathbf{h}}(\mathbf{h}^*, \mathbf{u}^*; \mathbf{\theta})\left(\mathbf{h}_{t-1} - \mathbf{h}^*\right) + \frac{\partial\mathbf{F}}{\partial \mathbf{u}}(\mathbf{h}^*, \mathbf{u}^*; \mathbf{\theta})\left(\mathbf{u}_t-\mathbf{u}^*\right)  \label{fo_apprx}.
\end{align}
Eq. (\ref{fo_apprx}) gives a LDS that locally approximates $\mathbf{F}$ around $\mathbf{h}^*$.  In principle, if one knew how to select the correct fixed point and there was always a fixed point nearby, one could use eq. (\ref{fo_apprx}) to run a very accurate SLDS approximation of $\mathbf{F}$ by switching between fixed points as needed.  Notice the fixed point $\mathbf{{h}}^*$ indexes the recurrent Jacobian and input Jacobian in a manner analogous to how the discrete state $z_t$ indexes the matrices  $A^{(z_t)}$ and $V^{(z_t)}$ in the SLDS of eq. (\ref{slds_eq}). In practice, selecting the correct fixed point in real-time is difficult, as described above, which leads to the approach presented in the next section.

\subsection{The JSLDS model}
Our approach is to co-train the RNN, $\mathbf{F}$, with a novel SLDS formulation based on the Jacobian of $\mathbf{F}$ in the spirit of equation (\ref{fo_apprx}).  Specifically, we introduce a separate SLDS with its own hidden state $\mathbf{a}_t \in \mathbb{R}^D$ that switches around an expansion point $\mathbf{e}_t^* \in \mathbb{R}^D$
\begin{align}
    \mathbf{a}_t - \mathbf{e}_t^* &= \frac{\partial\mathbf{F}}{\partial \mathbf{h}}(\mathbf{e}_t^*, \mathbf{u}^*; \mathbf{\theta})\left(\mathbf{a}_{t-1} - \mathbf{e}_t^*\right) + \frac{\partial\mathbf{F}}{\partial \mathbf{u}}(\mathbf{e}_t^*, \mathbf{u}^*; \mathbf{\theta})\left(\mathbf{u}_t-\mathbf{u}^*\right) \label{jac_slds}.
\end{align}
Note that eq. (\ref{jac_slds}) shares its parameters $\theta$ with $\mathbf{F}$ (eq. \ref{fotse}), i.e. given $\mathbf{e}_t^*$ and $\mathbf{u}^*$, the nonlinear RNN's parameters $\theta$ determine the update matrices. We will generally take $\mathbf{u}^*$ to be either zero (the average value for examples) or the value of a context-dependent static input.  The goal is for $\mathbf{e}_t^*$ to approximate the RNN's fixed and slow points.  To accomplish this, we supplement the SLDS with a nonlinear auxiliary function $\mathbf{E}$ (the \textbf{expansion network}) with separate learned parameters $\mathbf{\phi}$ 
\begin{align}
    \mathbf{e}_t^* &= \mathbf{E}(\mathbf{a}_{t-1}; \mathbf{\phi}) \label{expnet}.
\end{align}
The expansion network $\mathbf{E}$ returns the learned expansion points and is co-trained with the nonlinear RNN and the JSLDS.  Once trained, the goal is for this function to approximate the RNN's fixed/slow points. For the experiments presented in this paper, we define $\mathbf{E}$ as a 2-layer multilayer perceptron (MLP) with the same dimension per layer as the state dimension of the RNN. We discuss other potential formulations for this network in Section \ref{app_exp_net} in the Appendix.

Eqs. (\ref{jac_slds}-\ref{expnet}) define the JSLDS model, which can be run forward in time independent of the original nonlinear RNN after training. Figure \ref{graph_model} illustrates the general idea. Given the previous state, $\mathbf{a}_{t-1}$, the expansion network, $\mathbf{E}$, uses this state to select the next expansion point, $\mathbf{e}_t^*$ (eq. \ref{expnet}). The system then updates the state by using $\mathbf{e}_t^*$ to compute the recurrent Jacobian and input Jacobian of the original nonlinear RNN $\mathbf{F}$ (eq. \ref{jac_slds}). Assuming the expansion network has learned to find fixed/slow points, switching between the points $\mathbf{e}_t^*$ corresponds to reverse engineering nonlinear RNNs using fixed points. The dependence of the expansion point $\mathbf{e}_t^*$ on the previous state of the network $\mathbf{a}_{t-1}$ links to the recurrent connection in RSLDS. 

% of the same form as equation (\ref{fo_apprx}) and shares parameters $\mathbf{\theta}$ with the original nonlinear RNN $\mathbf{F}()$ and 

%Taken together, equations (\ref{jac_slds}-\ref{expnet}) form a closed loop system that can be run forward in time independent of the original nonlinear RNN in eq. (\ref{nlrnn}). 

% and the co-trained RNN as the RNN-J (to distinguish from a standard trained RNN).

Note that eq. (\ref{jac_slds}) is intended to closely follow the dynamics of $\mathbf{F}$, and we will enforce this in the training procedure. We will refer to the system comprised of eqs. (\ref{jac_slds}-\ref{expnet}) as the JSLDS. We will refer to the combination of the JSLDS and the co-trained RNN as the JSLDS-RNN system. While a limitation of our method is that it does not currently lend itself easily to a stochastic formulation like the standard SLDS, it does allow for a potential continuum of different switches using a constant number of parameters. It also automatically determines the number of switches required to solve the task instead of the hyperparameter tuning required to determine this in SLDS.

\subsection{JSLDS co-training Procedure}
We co-train together the nonlinear RNN (eq. \ref{fotse}) and the JSLDS (eqs. \ref{jac_slds}-\ref{expnet}).  Each network can be run forward and solve the task independently. We pass each of their states through the same output activation function to compute two loss functions, $\mathcal{L}_{\mathsf{RNN}}$ and $\mathcal{L}_{\mathsf{JSLDS}}$, for the RNN and JSLDS, respectively. In addition, the expansion points should approximate fixed points of $\mathbf{F}$ in order to achieve a good JSLDS approximation of the RNN. We also need to ensure the JSLDS states $\mathbf{a}_t$ approximate the RNN states $\mathbf{h}_t$. We achieve these goals by adding to the total loss function a fixed point regularizer $R_e$ and an approximation regularizer $R_a$ defined as 
\begin{align}
    R_e(\theta, \phi) &= \sum_t \left\|\mathbf{e}_t^* - \mathbf{F}(\mathbf{e}_t^*, \mathbf{u}^*; \mathbf{\theta})  \right\|_2^2 \label{reg_exp}\\
    R_a(\theta, \phi) &= \sum_t \left\|\mathbf{a}_t - \mathbf{h}_t\right\|_2^2 \label{reg_approx}.
\end{align}
Now we define the total training loss as 
\begin{align}
        \mathcal{L}(\theta, \phi) = 
        \lambda_{\mathsf{RNN}} \mathcal{L}_{\mathsf{RNN}}(\theta) 
        +\lambda_{\mathsf{JSLDS}} \mathcal{L}_{\mathsf{JSLDS}}(\theta, \phi)
        +\lambda_e R_e(\theta, \phi)
        +\lambda_a R_a(\theta, \phi) \label{JSLDS_loss}
\end{align}
where $\lambda_{\mathsf{RNN}}, \lambda_{\mathsf{JSLDS}}, \lambda_e$ and $\lambda_a$ control the strengths of the RNN loss, the JSLDS loss, the fixed point regularizer and the approximation regularizer, respectively. In practice, we have found these hyperparameters straightforward to select (see Section \ref{hyp_select}  in the Appendix for a more detailed discussion).  For a particular optimization iteration, we compute the loss function in eq. \ref{JSLDS_loss} and then update all of the parameters $\mathbf{\theta}$ and $\mathbf{\phi}$ at once using standard backpropagation through time (BPTT) methods for RNNs~\cite{werbos:bptt}. Assuming the optimization goes well, the result will be two independent trained systems, with the JSLDS approximating the nonlinear RNN to first order.

In related work,~\cite{schmidt2021identifying} introduced regularization terms that force part of the subspace of piecewise linear RNNs~\cite{koppe2019identifying} towards plane attractors to mitigate the exploding/vanishing gradient problem~\cite{hochreiter1997long, bengio1994learning} within a simple RNN architecture. In another relevant work,~\cite{duncker2019learning} proposed learning interpretable nonlinear SDEs by modeling the dynamics function as a Gaussian process conditioned on the learned locations of fixed points and associated local Jacobians. We also note that our co-training procedure shares some similarities to the adversarial training in GANs~\cite{NIPS2014_5ca3e9b1}. However, we stress that our method shares $\mathbf{\theta}$ between the co-trained networks, and $\mathbf{\theta}$ and $\mathbf{\phi}$ are each updated at the same time, i.e., we do not alternate between updating $\mathbf{\theta}$ holding $\mathbf{\phi}$ constant and vice versa.

\section{Results}

We analyze the JSLDS-RNN system on three examples: a synthetic 3-bit memory task, a synthetic context-dependent integration task, and multineuronal population recordings from a monkey performing a reaching task.\footnote{Our implementation for the synthetic tasks is available at https://github.com/jimmysmith1919/JSLDS\_public.} For the synthetic tasks, we use a relative error metric as in~\cite{NEURIPS2019_d921c3c7} to compare the quality of linearized approximation provided by JSLDS and the standard method of linearizing around the fixed/slow points (found numerically) of a standard trained RNN (without JSLDS co-training).  The metric computes the relative error, $\|\mathbf{h}_{t}^{\textsf{RNN}}-\mathbf{h}_t^{\textsf{lin}} \|_2/\|\mathbf{h}_t^{\textsf{RNN}} \|_2$ , between the nonlinear RNN state, $\mathbf{h}_t^{\textsf{RNN}}$, and the state approximated from the linearization method, $\mathbf{h}_t^{\textsf{lin}}$.  Note that for the standard linearization method we had to resort to only computing one-step ahead dynamics predictions.  This approach was necessary because running the linearized dynamics forward for many timesteps accumulates substantial error and causes the trajectory to diverge.  In contrast, one-step ahead predictions were not necessary for the JSLDS.

Concretely, for the standard linearization method: we trained an RNN, numerically found its fixed/slow points, and then computed $\mathbf{h}_t^{\textsf{lin}}$ for all of the timesteps of the held-out trials using eq. (\ref{fo_apprx}). To use eq. (\ref{fo_apprx}), we set $\mathbf{h_{t-1}}$ in that equation to $\mathbf{h}_{t-1}^{\textsf{RNN}}$, i.e., the true previous state and we set $\mathbf{h^*}$ to be the nearest fixed/slow point (in Euclidean distance).  We have to find the nearest fixed/slow point because, unlike JSLDS, the standard method does not directly link a location in state space to the fixed point one should linearize around. For the JSLDS method: we first co-trained the JSLDS and RNN together. We then simulated the JSLDS forward for the entire trajectory of each held-out trial using eqs. (\ref{jac_slds}-\ref{expnet}) to compute all of the JSLDS states $\mathbf{a}_t$. We then set  $\mathbf{h}_t^{\textsf{lin}}$ equal to $\mathbf{a}_t$ to compute the relative error for each timestep. For each method, we computed the mean relative error of all the timesteps of a held-out batch of trials.  We repeated this experiment 10 times for each method by starting the training from different random weight initializations. We report the mean and standard deviation of the mean relative error across the 10 trials.

\subsection{3-bit discrete memory}
\label{3bit_section}
\begin{figure}[t]
  \centering
  \includegraphics[width=1.0\linewidth]{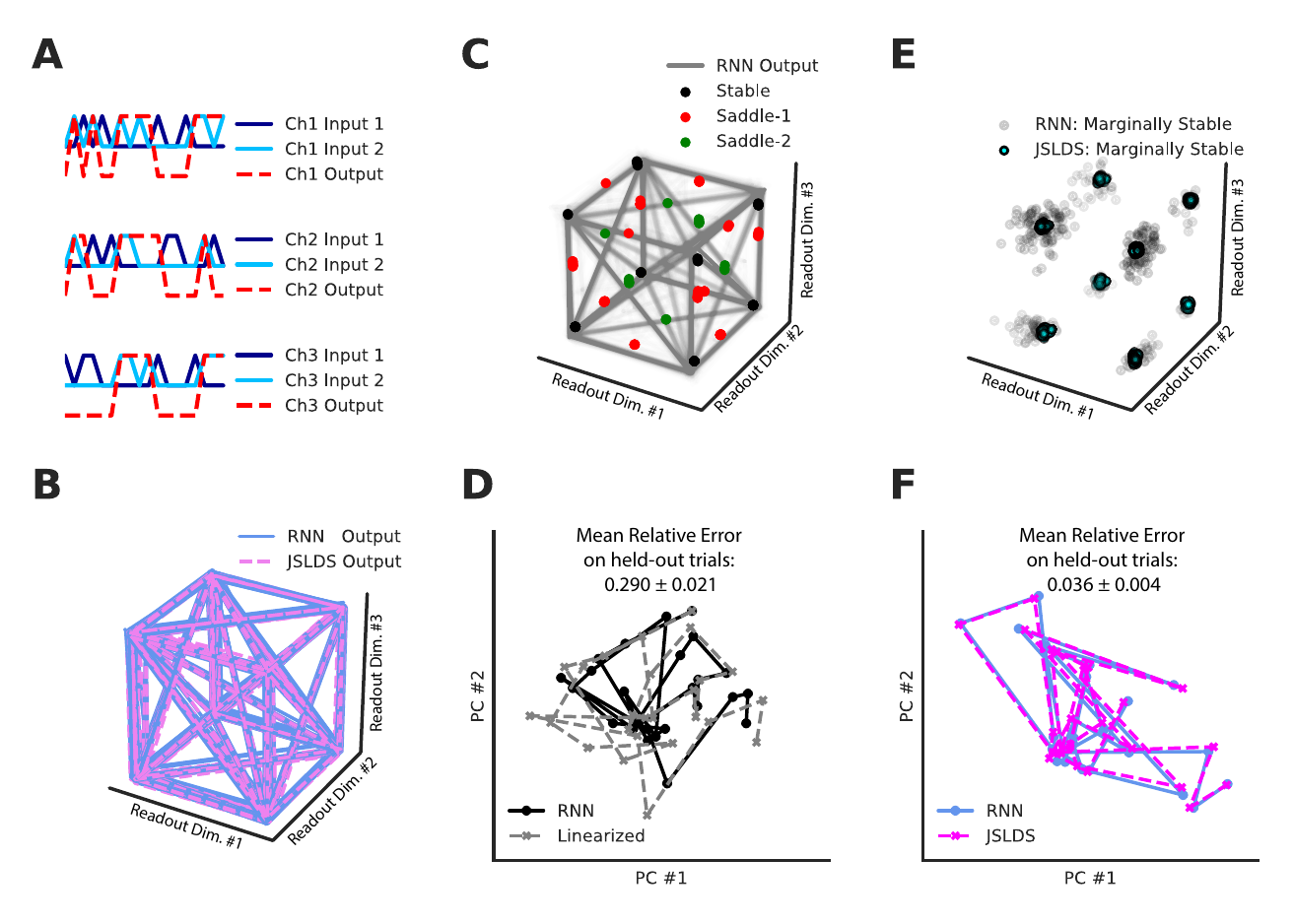}
  \caption{3-bit memory. \textbf{A.} Two-dimensional inputs (dark blue and light blue) corresponding to input states of -1, 0, or 1 enter at random while the corresponding output (dashed red) has to remember the last non-zero input state. \textbf{B.} JSLDS closely approximates the co-trained RNN in readout space for held-out trial data. \textbf{C.} Standard GRU (no JSLDS co-training) outputs and numerical fixed points projected into readout space.   \textbf{D}. Example PCA trajectory of standard GRU and linearized dynamics (one-step ahead dynamics generation) using numerically optimized fixed points. We also note the mean relative error for held-out trials. \textbf{E}. Comparison of co-trained RNN fixed points (found numerically) and JSLDS expansion points projected into readout space. The solution consists of 8 marginally stable fixed points. JSLDS has changed the fixed point solution compared to the standard GRU's solution (panel C). \textbf{F}. Example PCA trajectory of co-trained RNN and JSLDS dynamics (fully simulating dynamics forward) and the held-out mean relative error.}
  \label{3bit_fig}
\end{figure}

This task highlights how JSLDS can automatically learn to switch about a discrete number of fixed points and significantly reduces the linearized approximation error. We trained the JSLDS-RNN system to store and output three discrete binary inputs (Figure \ref{3bit_fig}A) similar to the experiment described in~\cite{Sussillo_Barak}. For our purposes, the models receive three 2-dimensional input vectors where each input vector corresponds to a different channel.\footnote{We reparameterized the inputs compared to~\cite{Sussillo_Barak}. This ensures the models do not have to act nonlinearly in the inputs and does not change the basic logic of the experiment since we are interested in nonlinear dynamics.} Each input vector can take a value of $\{[1,0], [0,0], [0,1]\}$ corresponding to a state of -1, 0, or +1, respectively. The models have three outputs, each of which needs to remember the last nonzero state of its corresponding input channel. The RNNs used in this experiment were GRUs~\cite{cho-etal-2014-learning} with a state dimension of 100 and a linear readout function. See Section \ref{3bit_methods} of the Appendix for additional experiment details.

Projecting the co-trained RNN and JSLDS dynamics into the readout space illustrates the close agreement between the networks for predictions on held-out trials (Fig. \ref{3bit_fig}B). A benefit of JSLDS is that we can use its expansion points produced along a trajectory (given the trajectory inputs) to approximate the fixed/slow points the co-trained RNN uses along the same trajectory.  As a verification, we numerically found the fixed/slow points of the co-trained RNN and projected both the expansion points and the numerical fixed/slow points into the readout space (Fig. \ref{3bit_fig}E). We observed that the co-trained networks learned a fixed point solution consisting of 8 marginally stable fixed points (typically 2-3 eigenvalues within .025 of (1,0) in the complex plane). This solution was robust across different random weight initializations.  See Section \ref{app_fps_sol} of the Appendix for a detailed analysis and discussion of this solution compared to the fixed point solution found by a standard GRU without co-training (Fig. \ref{3bit_fig}C).  Figures \ref{3bit_fig}D and F compare the mean relative error for the standard method and the JSLDS and show an example PCA trajectory of the reconstructed dynamics for each. This shows that JSLDS can simulate forward entire dynamics trajectories with much less error than the standard method (which relies on one-step ahead dynamics generation).  Finally, as an additional experiment, we initialized the JSLDS co-training procedure with the trained weights of the standard GRU. We observed the fixed point solution change from the one presented in Fig. \ref{3bit_fig}C to one like that presented in Fig. \ref{3bit_fig}E and the same improved linearized approximations of the dynamics as presented in Fig. \ref{3bit_fig}F.  

\subsection{Context-dependent Integration}
\begin{figure}[ht!]
  \centering
  \includegraphics[width=\linewidth]{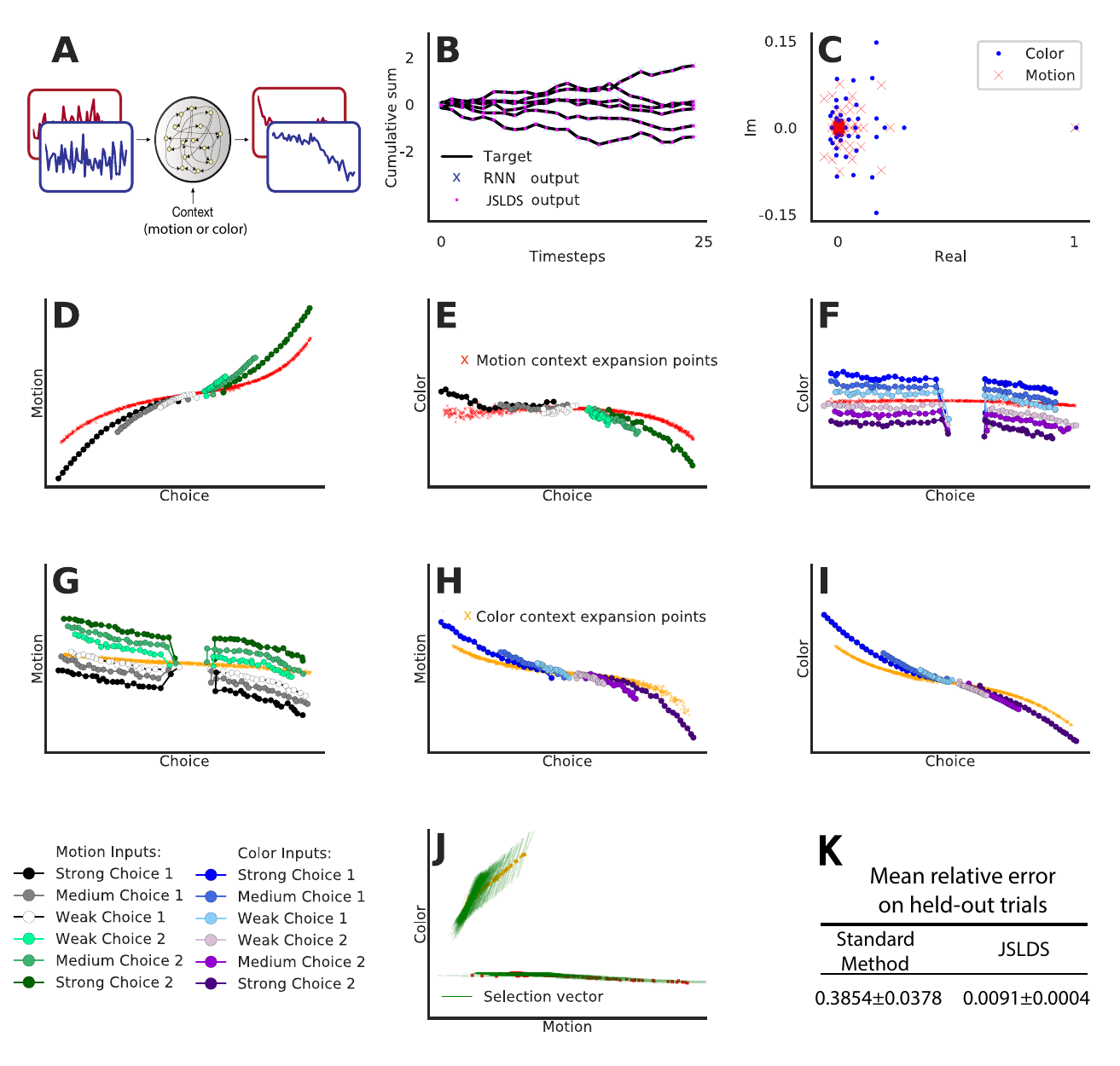}
  \vspace{-1em}
  \caption{Context-dependent integration \textbf{A.} One of two white-noise input streams (motion or color) is  selected to be integrated based on a static context input. The other stream is ignored. \textbf{B.} Sample held-out trial outputs show close agreement between JSLDS and RNN. \textbf{C.} Typical eigenvalues at a sample expansion point for motion (red x's) and color (blue dots) contexts.  \textbf{D-J.} JSLDS has learned to switch between two continuous manifolds of fixed points. JSLDS states (averaged) and expansion points are projected into the subspace spanned by the axes of choice, motion and color. Movement along the choice axis represents integration of evidence and the relevant input stream deflects along the relevant input axis. The input axes of \textbf{E,F,G,H} have been intensified. The trials used in \textbf{F} and \textbf{G} are the same trials as \textbf{D-E} and \textbf{H-I}, respectively, but re-sorted and averaged according to the direction and strength of the irrelevant input. The expansion points were computed separately for motion (red x's) and color contexts (orange x's). \textbf{J.} Global arrangement of the selection vectors (green lines) and line attractor expansion points for both contexts projected onto the input axes. Inputs are selected by the selection vector (which is approximately orthogonal to the contextually irrelevant input) and integrated along the line attractor. \textbf{K.} JSLDS improves the dynamics approximation compared to linearizing a standard trained RNN.}
  \label{context_int_fig}
\end{figure}

This task illustrates that JSLDS can learn to switch about multiple continuous manifolds of fixed points, improve the linearized approximation of the dynamics, and be used to perform a complex analysis similar to that performed in~\cite{mante2013context}.  The experiment consists of training the model to contextually decide which of two white noise input streams to integrate (Fig. \ref{context_int_fig}A). The model receives two static context inputs corresponding to motion and color contexts and two time-varying white noise input streams. It is trained to output the cumulative sum of the white noise stream specified by the active context input. We used a vanilla RNN with a state dimension of 128 for the co-trained RNN and a linear readout function. See Section \ref{contextint_methods} of the Appendix for more experiment details.

After co-training the JSLDS-RNN system, we observed close agreement between the JSLDS and RNN on task performance for held-out trials (Fig. \ref{context_int_fig}B). Next, we analyzed the dynamics of the JSLDS for held-out trials under both contexts, set to different bias levels for both the color and motion input streams.  We observed that for either context, the system integrates the relevant input using a single linear mode with an eigenvalue of 1, while the other eigenvalues decay rapidly (Fig. \ref{context_int_fig}C).  Next, we report the mean relative error on held-out trials for both JSLDS and linearizing a standard trained RNN in Fig. \ref{context_int_fig}K. Again, JSLDS substantially improves the linearized approximation of the dynamics. 

From the analysis in~\cite{mante2013context}, we expect a vanilla RNN to solve this type of task by representing the integration of relevant evidence as movement along an approximate line attractor (the \textbf{choice axis}) determined by the top right eigenvector. The solution consists of two line attractors that never exist together: one exists in the motion context and the other in the color context.  For a given context (and therefore a specific line attractor), the top left eigenvector (the \textbf{selection vector}) determines the amount of evidence integrated. We expect the selection vector to project strongly onto the relevant input and be approximately orthogonal to the irrelevant input. See the Mathematical Supplement Section 10 in~\cite{mante2013context} for more details.

To verify this holds for the JSLDS, we produce a figure similar to Figures 5 and 6c from~\cite{mante2013context} by projecting the JSLDS states and expansion points into the 3-dimensional subspace meant to match the axes of choice, motion input, and color input (Fig. \ref{context_int_fig}D-I). Section 7.6 of the Supplementary Information of~\cite{mante2013context} describes the construction of this subspace in detail and we provide a brief description in Section \ref{contextint_methods_subs} in the Appendix. It is analogous to a regression subspace estimated from neural data in that work. It was constructed by orthogonalizing the direction of the first right eigenvectors (averaged over expansion points) and the input weight vectors corresponding to the color and motion input streams. Panels D-F and G-I correspond to the motion and color contexts, respectively. In panels D and I, we see that for the relevant context input stream, the states move along the axis of choice and the relevant input axis in proportion to the strength of the input.  Panels F and G show that the strength of the nonrelevant input stream does not affect the direction of choice. Next,  we analyze the global arrangement (Fig. \ref{context_int_fig}J) of the motion and color context line attractors and selection vectors. As expected, we see that the selection vectors project strongly onto the relevant input axis but are approximately orthogonal to the irrelevant axis. Figure \ref{context_int_nojslds_fig} in the Appendix presents the results of performing this experiment for a standard trained vanilla RNN without the JSLDS co-training.  It appears the JSLDS co-training did not dramatically change the standard trained RNN's fixed point solution for this task.

% \begin{figure}[t]

\subsection{Monkey reach task with LFADS-JSLDS}
Finally, we illustrate how JSLDS can be dropped in as a module to improve our understanding of more complex architectures that use RNNs such as LFADS~\cite{pandarinath2018inferring}. LFADS is a sequential variational auto-encoder~\cite{kingma2013auto,rezende2014stochastic} used to infer latent dynamics from single-trial neural spiking data. A criticism of LFADS has been that it is hard to interpret the RNN generator that produces the dynamics. Here, we use JSLDS to improve this understanding by substituting the combined JSLDS-RNN system for the standard GRU used in the LFADS generator (Fig. \ref{lfads_jslds_fig} in the Appendix).  We refer to this system as the LFADS-JSLDS. Once trained, we can use either the JSLDS or the co-trained RNN as the generator to produce the firing rates.

We used the monkey J single-trial maze data from~\citet{churchland2012neural} using the same setup as~\citet{pandarinath2018inferring} to train the LFADS-JSLDS model. The data consists of 2296 trials of spiking activity recorded from 202 neurons simultaneously while a monkey made reaching movements during a maze task~\cite{churchland2010cortical,churchland2012neural} across 108 reaching conditions. We used a GRU for the RNN generator. See Section \ref{monkey_methods} of the Appendix for more details. The jPCA method ~\cite{churchland2012neural} has been applied to this data before ~\cite{sussillo2015neural, pandarinath2018inferring}, so we make use of it to validate our method. It finds linear combinations of principal components that capture rotational structure in data.  See~\cite{churchland2012neural} for the full details on jPCA.

\begin{figure}[h!]
  \centering
  \includegraphics[width=1.0\linewidth]{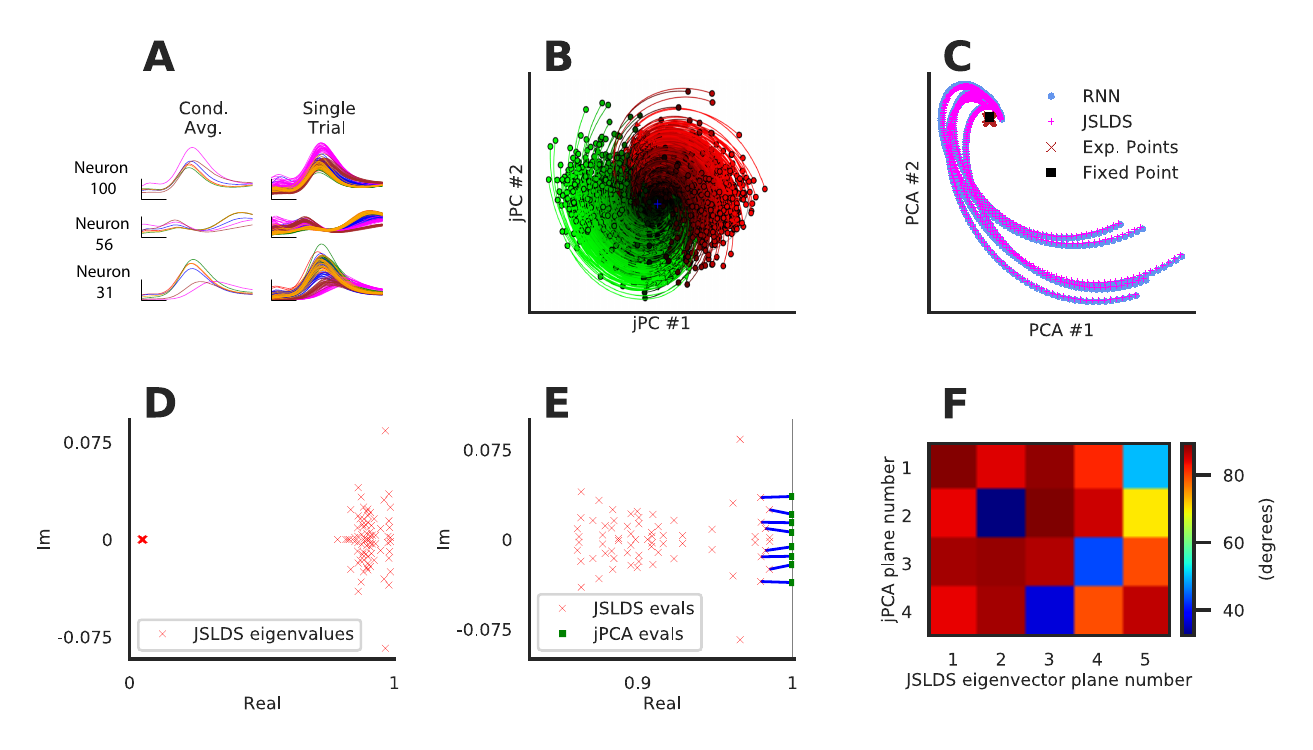}
  \caption{Monkey maze task. \textbf{A.}  LFADS-JSLDS firing rates generated from inferred initial conditions for sample neurons. \textbf{B}.~Projection of the co-trained RNN generator hidden states onto the first 2 jPC planes. We see the generator exhibits the expected rotational dynamics.  \textbf{C}. Sample trial dynamics show low approximation error between RNN and JSLDS. Note also the JSLDS expansion points and RNN numerical fixed point. \textbf{D}. Eigenvalues of the JSLDS at a single timestep of a single trial. Our observation is that these eigenvalues are the same for every timestep of every trial, i.e. LFADS-JSLDS has learned to organize the movement dynamics with a single condition-independent linear system. \textbf{E}. Top 70 eigenvalues from the same linear system shown in D along with the purely imaginary eigenvalues associated with the jPCA analysis (green squares). The jPCA eigenvalues are connected (blue line) to their corresponding JSLDS eigenvalues given by the subspace analysis in F.  \textbf{F}. Subspace analysis comparing the jPC planes and planes corresponding to the top five complex eigenvalue pairs of the JSLDS generator. Color indicates the minimum subspace angle between the corresponding planes. Angles of 30-40 degrees indicate highly overlapping subspaces.}
  \label{monkey_fig}
\end{figure}

We present the LFADS-JSLDS firing rates generated from the inferred initial condition (using the co-trained RNN generator) for several sample neurons in Fig. \ref{monkey_fig}A. It was observed in ~\cite{pandarinath2018inferring} that the standard LFADS population dynamics on single trials exhibit rotational dynamics when projected onto the first two jPC planes. To confirm LFADS-JSLDS also exhibits this behavior, we applied jPCA to the co-trained RNN generator states (Fig. \ref{monkey_fig}B). Next, the sample trials in Figure \ref{monkey_fig}C show how the JSLDS generator closely approximates the RNN generator. Finally, focusing on the JSLDS generator dynamics, we learned that despite minor variations in the expansion points, the JSLDS generator eigenvalues and eigenvectors were the same at every timestep of every trial. So the JSLDS generator learned to represent the dynamics using a single, condition-independent linear system. Figure \ref{monkey_fig}D displays the eigenvalues for this system.  

\citet{sussillo2015neural} noted that the dynamics found by linearizing around the fixed point of their regularized RNN model should roughly agree with the dynamics found by applying jPCA directly to the model. Along these lines, we validate the use of JSLDS by confirming that the JSLDS generator dynamics agree with the dynamics found by fitting jPCA to the co-trained RNN generator states (Fig. \ref{monkey_fig}E-F). A subspace angle analysis shows that four of the top five planes in the JSLDS state space (defined by the eigenvectors) were similar to the first four jPCA planes (Fig. \ref{monkey_fig}F). Using this correspondence, we see that the eigenvalues reported by jPCA (constrained to be purely imaginary) revealed four frequencies that closely agreed with the top four frequencies found by analyzing the JSLDS generator dynamics (Fig~\ref{monkey_fig}E). See Section \ref{monkey_methods_subspace} for more details on this subspace analysis. The correspondence between the jPCA and the JSLDS dynamics validates the use of JSLDS for the LFADS generator. Given this, we conclude that LFADS-JSLDS learns to represent the dynamics for this task using a single, condition-independent linear system, a fact that was not obvious a priori. See Section \ref{monkey_methods_no_jslds} of the appendix for a discussion of numerically finding the fixed points of a trained LFADS model without the JSLDS co-training. We observed that the JSLDS co-training did not dramatically change the standard LFADS model's fixed point solution for this task. 

\section{Discussion}
Inspired by ideas from reverse engineering RNNs~\cite{Sussillo_Barak, mante2013context,NEURIPS2019_d921c3c7,Maheswaranathan2020HowRN, sussillo2015neural,NEURIPS2019_5f5d4720,NEURIPS2020_30f0641c} and SLDS models~\cite{ pmlr-v54-linderman17a,Linderman621540,nassar2018treestructured, NEURIPS2020_aa1f5f73}, this work addresses the challenging problem of improving our understanding of how RNNs perform computations. We introduced a new model, the JSLDS, to improve our ability to reverse engineer RNNs. We applied it in various settings and to different architectures: GRUs, Vanilla RNNs, and LFADS models. The JSLDS does not require post-training fixed point optimizations, significantly reduces the approximation error associated with reconstructing the nonlinear dynamics using the locally linearized solutions, and maps each point in state space to a fixed point.  These benefits significantly improve our ability to reverse engineer RNNs, assuming a state-dependent SLDS can provide a good approximation of the original nonlinear RNN trained on a particular task. 

Furthermore, the JSLDS could be a valuable tool to investigate the limits of the general framework of reverse engineering RNNS with fixed points. This is because we only expect the JSLDS approximation to break down if there is a system with nonlinear dynamics that are not well-described by switching between linearizations around fixed points.  In addition to the above, JSLDS generalizes SLDS models to a potential continuum of switches with a constant number of parameters and automatically learns the required number of switches. 

An area for refinement is the expansion network. We observed in some experiments that the expansion network might produce clusters of slightly varying expansion points that all define a single linear system instead of just producing a single expansion point.  Perhaps an additional loss function penalty or more specific architectures for particular tasks could help reduce this variation.

Finally, in the 3-bit memory task, the JSLDS co-training changed the fixed point structure the co-trained RNN used to solve the task compared to the standard GRU solution. We additionally observed improved linearized dynamics approximations with this new solution. These observations provide evidence that JSLDS can regularize a nonlinear RNN towards solutions better described by switching between linearized dynamics around fixed points. This regularization towards a switching linear structure could potentially have beneficial performance effects for robustness and generalization on held-out data. However, more in-depth and larger-scale studies are required to quantify these potential effects. We also note that our method could potentially suffer from the same theoretical limitation discussed for the previous reverse engineering method in Section \ref{rev_eng} when near non-hyperbolic fixed points due to the Hartman-Grobman theorem. However, the potential for our method to bias the nonlinear RNN solutions towards solutions well approximated by switching between linearizations around fixed points could alleviate this concern in practice.

\textbf{Broader Impact} While understanding how RNNs perform complex computations could eventually help bound expected model behavior, identify biases, improve robustness to adversarial inputs and suggest ways to improve performance, we foresee no immediate societal consequences of this work.
Overall, this work makes it easier to reverse engineer RNNs and continue to gain insight into how these models work to benefit both neuroscience and machine learning.

\clearpage

\textbf{Acknowledgements} We would like to thank Mark Churchland, Matt Kaufman and Krishna V. Shenoy for access to the monkey maze data.

J.T.H.S. received funding support from a Stanford Graduate Fellowhip in Science and Engineering (Mayfield fellowship). S.W.L. was supported by grants from the Simons Collaboration on the Global Brain (SCGB 697092) and the NIH BRAIN Initiative (U19NS113201 and R01NS113119). D.S. was supported by a grant from the Simons Foundation (SCGB 543049, DCS).

\medskip

\bibliographystyle{unsrtnat}
\bibliography{refs}

\clearpage
\appendix
\section{Appendix}
\setcounter{figure}{0} 
\subsection{JSLDS hyperparameter selection}
\label{hyp_select}
In general, we have found the JSLDS loss function strengths to be relatively easy to select (see example settings in the specific experiment sections below). However, there are various possible configurations. The following provides a general framework for how to think about these parameters:
\begin{itemize}
    \item $\lambda_e$ should generally be relatively large. It should be prioritized higher than the other losses or regularizers since failing to find expansion points that are good approximations of the RNN's fixed points or slow points would defeat the primary purpose of the method.
    \item $\lambda_a$ should be large enough to ensure a small error between the JSLDS and RNN states. However, for some tasks, one may need to balance tradeoffs between $R_a$ and the losses $\mathcal{L}_{\mathsf{RNN}}$ and $\mathcal{L}_{\mathsf{JSLDS}}$.
    \item For most of the experiments in this paper we set the loss strengths to $\lambda_{\mathsf{RNN}} = 1$ and $\lambda_{\mathsf{JSLDS}}=1$. For the 3-bit memory task, we observed slightly better performance by setting $\lambda_{\mathsf{RNN}} = 3$ and $\lambda_{\mathsf{JSLDS}}=1$. Interestingly, allowing for a slight bias towards the RNN performance on this task generally led to improved performance for both the RNN and the JSLDS. However, other variations are possible.
    \item For example, setting $\lambda_{\mathsf{RNN}} = 1$ and $\lambda_{\mathsf{JSLDS}}=0$ might correspond to the goal of training a nonlinear RNN to be more interpretable by not sacrificing the goals of $R_e$ and $R_a$ for the sake of JSLDS task performance. 
    \item In the other extreme, if one were just interested in training an SLDS, setting $\lambda_{\mathsf{RNN}} = 0$ and $\lambda_{\mathsf{JSLDS}}=1$ could provide benefits since the JSLDS learns to share parameters across expansion points.
\end{itemize}

\subsection{Expansion network formulation}
\label{app_exp_net}
In this work we considered a specific formulation of the expansion network as $\mathbf{E}(\mathbf{a}_{t-1}; \mathbf{\phi})$, in which the network is a 2-layer MLP that only depends on the previous state. However other formulations are possible. For example, the expansion network could also depend on the previous expansion point $\mathbf{E}(\mathbf{a}_{t-1}, \mathbf{e}_t^*; \mathbf{\phi})$ , the input $\mathbf{E}(\mathbf{a}_{t-1}, \mathbf{u}_t; \mathbf{\phi})$, or a combination of all of these $\mathbf{E}(\mathbf{a}_{t-1}, \mathbf{e}_t^*,\mathbf{u}_t; \mathbf{\phi})$. It is interesting future work to study the effects of variations such as these.

\subsection{3-bit discrete memory task}
\label{3bit_methods}
\subsubsection{Experimental details}
The task here consists of three 2-dimensional input vectors where each input vector corresponds to a different channel. Each input vector can take a value of $\{[1,0], [0,0], [0,1]\}$ corresponding to a state of -1, 0, or +1 respectively. The models have three outputs, each trained to remember the last nonzero state of its corresponding input channel.   For example, output 2 remembers whether channel 2 was last set to state +1 or -1, but ignores the channel 1 and channel 3 inputs.  When a given channel receives a nonzero input that is different from its current state, it should immediately output the new state on that timestep.  For a given input vector at a given timestep, we set the probability of being in any of the three states to be equal. We set the number of timesteps $T=25$.

We trained both methods with the Adam optimizer with default settings. For the JSLDS, we set the value of $\mathbf{u}^*$ in the JSLDS to zero. Other important hyperparameters are listed in Table \ref{3bittable}.

\begin{table}[h]
  \caption{Hyperparameters used for 3-bit memory task}
  \label{3bittable}
  \centering
  \begin{tabular}{lll}
    \toprule
    \cmidrule(r){1-2}
    Model     & JSLDS-RNN     & Standard RNN \\
    \midrule
    RNN type & GRU  & GRU     \\
    Number of RNN layers          & 1 & 1 \\
    Hidden state dimension        & 100 & 100 \\
    Batch size                    & 256 &256 \\
    Initial learning rate         & .02 &  .02 \\
    L2 regularization             &0.0  & 0.0\\
    Expansion network layers      & 2   & n/a   \\
    Expansion network units/layer & 100 & n/a   \\
    Expansion network activation  & tanh & n/a \\
    $\lambda_{\mathsf{RNN}}$                & 3.0 & n/a \\
    $\lambda_{\mathsf{JSLDS}}$              & 1.0 & n/a \\
    $\lambda_{e}$                 & 100.0 & n/a \\
    $\lambda_{a}$                 & 10.0 & n/a \\
    \bottomrule
  \end{tabular}
\end{table}

\subsubsection{JSLDS co-training fixed point solution}
\label{app_fps_sol}
The fixed point solution used by the JSLDS and co-trained GRU to solve the 3-bit memory task is significantly different from the solution used by the standard GRU (without co-training). As displayed in the main paper, the standard fixed point solution consists of stable fixed points on the corners and saddle nodes in between. In contrast, the JSLDS co-training results instead in a solution that consists of only marginally stable fixed points.  We note there does seem to be some variability in the expansion network that causes the expansion points to form clusters instead of distinct points. However, this variability is relatively small in the sense that within any of the distinct clusters, all the expansion points have nearly identical linearizations. This is confirmed by checking the eigenvalues and eigenvectors for the points within each cluster. Therefore, the eight distinct clusters of marginally stable expansion points define what is essentially eight marginally stable fixed points for each of the eight possible target output states.

As we presented, the JSLDS only utilizes these eight marginally stable points. In addition, when using the numerical fixed point finding method, the slowest of the numerical fixed points of the co-trained RNN also cluster around these eight points and are also marginally stable. We can also adjust the tolerance threshold used by the numerical fixed point finding method to observe less slow points. This reveals more marginally stable fixed points in between the eight corners. These marginally stable points between the corners stand in contrast to the saddle nodes present between the corners in the standard solution. As we also noted, initializing the JSLDS co-training procedure with the trained weights of the standard GRU also leads to this same marginally stable solution. I.e., the JSLDS co-training changes the fixed point solution from the classic solution to our new solution with marginally stable fixed points. These results suggest that perhaps one can think of the JSLDS co-training as acting to make the stable corners of the standard solution less stable and the unstable saddles of the standard solution more stable, resulting in the marginally stable solution we observe.

The co-trained JSLDS-RNN solution uses these eight marginally stable fixed points to dynamically select or ignore the inputs to update the hidden state. This is made apparent by studying the top left eigenvectors of the recurrent Jacobian (which we will refer to as the selection vectors) at each of the eight clusters and how they act upon the different effective inputs.  Recall the input vector for each of the three channels can take a value of $\{[1,0], [0,0], [0,1]\}$. Because the $[0,0]$ input will have no effect, we can focus on just six inputs corresponding to the six one-hot input vectors that could flip one of the channel output states. We can view these six inputs for $\mathbf{u}_t$ as a $6\times6$ identity matrix where each column represents a different input that we are interested in. The effective input for the JSLDS update equation is $\frac{\partial\mathbf{F}}{\partial \mathbf{u}}(\mathbf{e}^*, \mathbf{u}^*; \mathbf{\theta})\left(\mathbf{u}_t-\mathbf{u}^*\right)$. We can use our identity matrix as the different $\mathbf{u}_t$'s, which allows us to represent the different effective inputs we are interested in as a  $100\times6$ matrix (where 100 is the hidden state dimension used in this experiment).

\begin{figure}[ht!]
  \centering
  \includegraphics[width=1.0\linewidth]{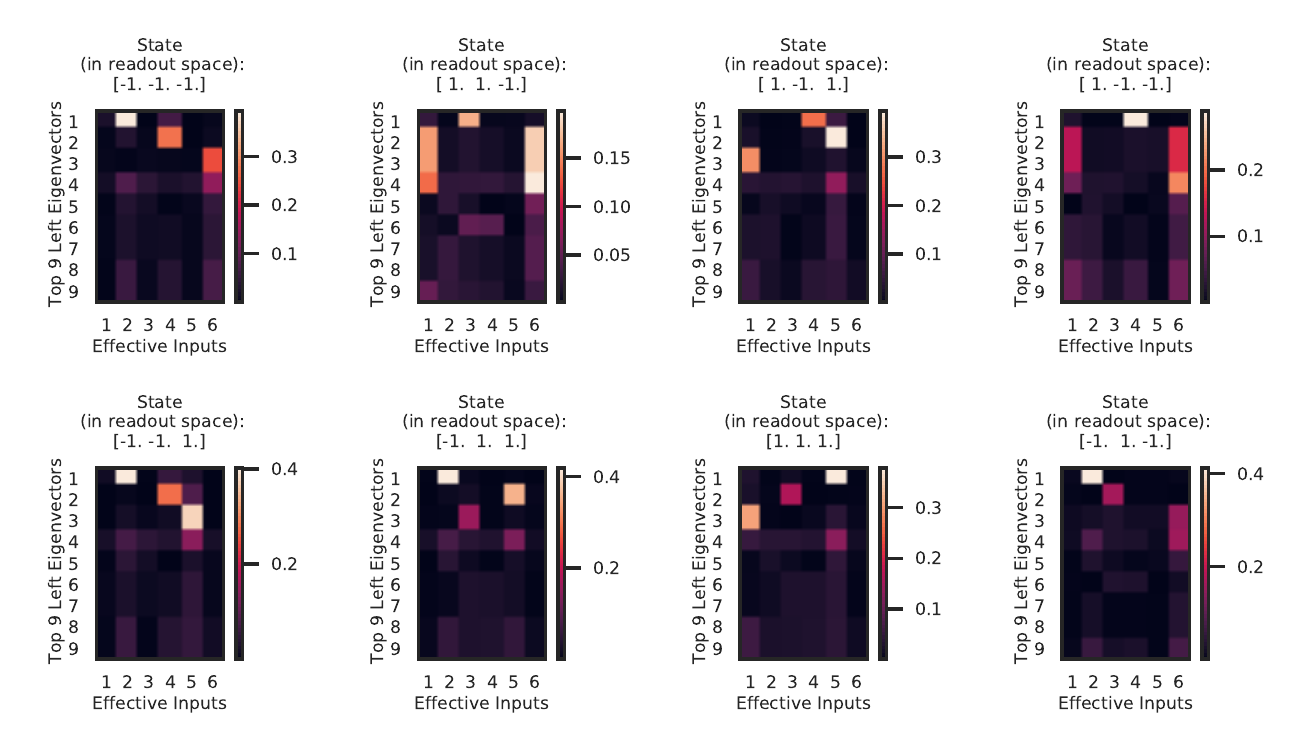}
  \caption{Analysis of the hidden state update mechanism for the 3-bit memory task. For each of the eight expansion points the JSLDS solution uses for the eight possible output states, we take the dot product between the top nine left eigenvectors of the recurrent Jacobian $\frac{\partial\mathbf{F}}{\partial \mathbf{h}}(\mathbf{e}^*, \mathbf{u}^*; \mathbf{\theta})$, represented as a $9\times100$ matrix, and the effective input $\frac{\partial\mathbf{F}}{\partial \mathbf{u}}(\mathbf{e}^*, \mathbf{u}^*; \mathbf{\theta})\left(\mathbf{u}_t-\mathbf{u}^*\right)$ for each of the six one-hot inputs $\mathbf{u_t}$ we are interested in,  represented as a $100\times6$ matrix. This results in a $9\times6$ matrix for each of the eight possible output states. Note that both $\frac{\partial\mathbf{F}}{\partial \mathbf{h}}(\mathbf{e}^*, \mathbf{u}^*; \mathbf{\theta})$ and $\frac{\partial\mathbf{F}}{\partial \mathbf{u}}(\mathbf{e}^*, \mathbf{u}^*; \mathbf{\theta})$ depend on the expansion point.  The resulting dot product values have been normalized. We see that essentially the only nonzero results correspond to the inputs that would flip the corresponding channel state. For example, in the bottom right, state [-1,1,-1], the nonzero dot products correspond to the second, third, and sixth effective inputs. This corresponds to actual inputs of $[0,1], [1,0], [0,1]$ for each of the three channels respectively. According to the task definition, these are the inputs that would cause each of the three channels to flip its respective output state.  On the other hand, the first, fourth, and fifth effective inputs have no effect. This corresponds to actual inputs of  $[1,0], [0,1], [1,0]$ for the three channels respectively. According to the task definition, these inputs should not impact the states, as we observe.}
  \label{leftevec_dot_fig}
\end{figure}

We can take the dot product between the selection vectors and the different effective inputs to reveal visually intuitive patterns that clarify how the selection vectors correctly select or ignore the inputs when the system is in a particular state. We can view the top 9 left eigenvectors as a $9\times100$ matrix and multiply this by our $100\times6$ effective input matrix. Figure \ref{leftevec_dot_fig} presents the results, and we observe that the normalized dot product between the selection vectors and the effective inputs is essentially only nonzero for the effective inputs that would cause one of the channel output states to flip. 

\begin{figure}[h!]
  \centering
  \includegraphics[width=1.0\linewidth]{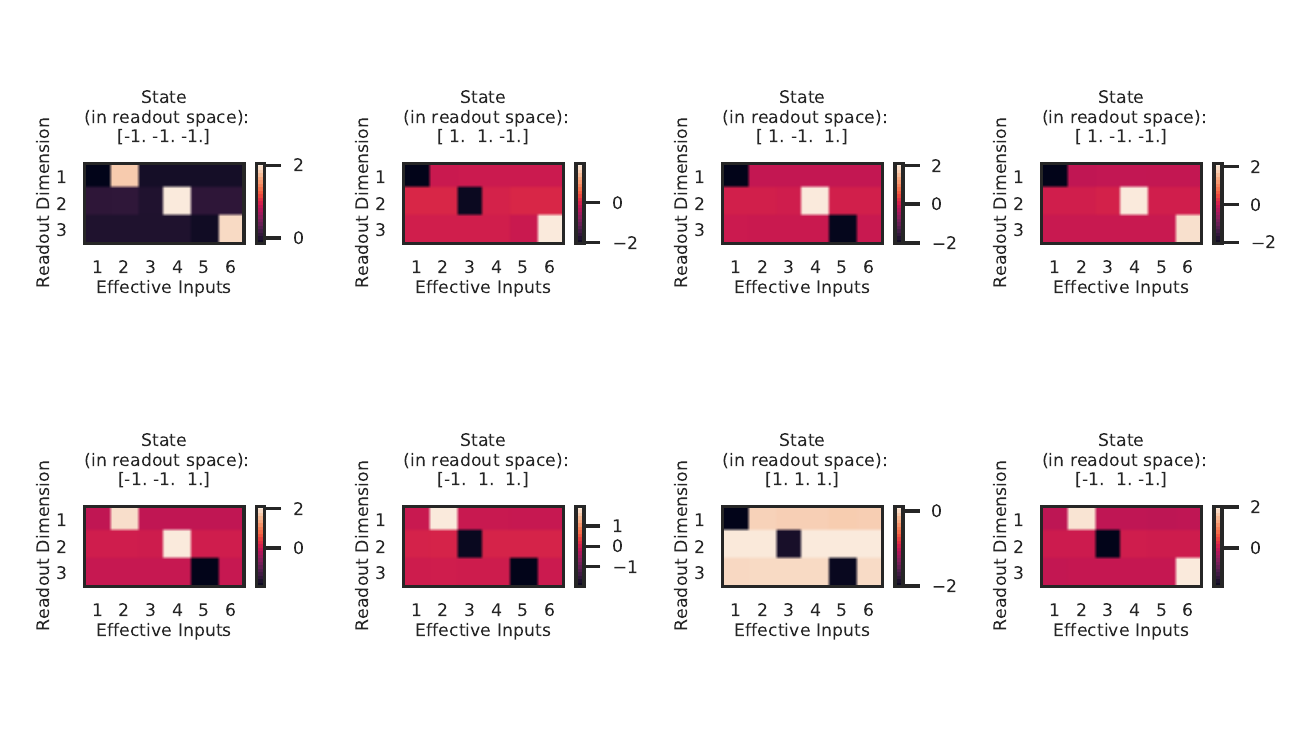}
  \caption{Analysis of the readout state update mechanism for the 3-bit memory task. For each of the eight expansion points the JSLDS solution uses for the eight possible output states, we take the dot product between the readout matrix, represented as a $3\times100$ matrix, and the effective input $\frac{\partial\mathbf{F}}{\partial \mathbf{u}}(\mathbf{e}^*, \mathbf{u}^*; \mathbf{\theta})\left(\mathbf{u}_t-\mathbf{u}^*\right)$ for each of the six one-hot inputs $\mathbf{u_t}$ we are interested in,  represented as a $100\times6$ matrix. This results in a $3\times6$ matrix for each of the eight possible output states. We see that essentially the only nonzero results correspond to the inputs that would flip the corresponding channel output state. }
  \label{readout_dot_fig}
\end{figure}

For this task, the system should immediately update a channel output state on the same timestep it receives a nonzero input that is different from the current channel state. We can observe how the system performs this update by taking the dot product between the different possible effective inputs and the readout matrix (a size $3\times100$ matrix). This is because the readout matrix must use the effective input to make this update immediately. Figure \ref{readout_dot_fig} presents the results.  We see that the dot product is essentially only nonzero for the dimensions corresponding to the effective inputs that would cause the corresponding channel output state to flip.

\clearpage

\subsection{Contextual integration task}
\subsubsection{Experimental details}
\label{contextint_methods}

The experiment consists of training vanilla RNNs to contextually decide which of two white noise input streams, corresponding to motion or color contexts, to integrate. Models received two static context inputs, corresponding to motion and color contexts, and two time-varying white noise input streams of length $T=25$. On each trial, one context input was zero and the other one, forming a one-hot encoding that indicates which input stream should be integrated. The white noise input was sampled from $\mathcal{N}(\mu, .125^2)$ at each time step, with $\mu$ sampled from $\mathcal{N}(-.01,.02^2 )$ and kept static across time for each trial. The models were trained to output the cumulative sum of the correct context white noise stream at each timestep. For evaluation, the fixed $\mu$s used for the inputs were [-.04,-.02,-.009,.009,.02,.04] corresponding to strong, intermediate and weak evidence for both choices.

For $\mathbf{u}^*$ in the JSLDS, we set the dimensions that correspond to the white noise inputs to zero and set the other dimensions to the value of the context-dependent static input for each trial.  We trained the system using the Adam optimizer with default settings. Other important hyperparameter settings are listed in Table \ref{contextinttable}.

\begin{table}[h!]
  \caption{Hyperparameters used for contextual integration task}
  \label{contextinttable}
  \centering
  \begin{tabular}{lll}
    \toprule
    \cmidrule(r){1-2}
    Model     & JSLDS-RNN     & Standard RNN \\
    \midrule
    RNN type & Vanilla  & Vanilla     \\
    Number of RNN layers          & 1 & 1 \\
    Hidden state dimension        & 128 & 128 \\
    Batch size                    & 256 &256 \\
    Initial learning rate         & .02 &  .02 \\
    L2 regularization             &1.0e-5  & 1.0e-5\\
    Expansion network layers      & 2   & n/a   \\
    Expansion network units/layer & 128 & n/a   \\
    Expansion network activation  & tanh & n/a \\
    $\lambda_{\mathsf{RNN}}$                & 1.0 & n/a \\
    $\lambda_{\mathsf{JSLDS}}$              & 1.0 & n/a \\
    $\lambda_{e}$                 & 100.0 & n/a \\
    $\lambda_{a}$                 & 10.0 & n/a \\
    \bottomrule
  \end{tabular}
\end{table}

\subsubsection{Subspace construction}
\label{contextint_methods_subs}
To display the RNN trajectories in state space, we projected the JSLDS states and expansion points into the 3-dimensional subspace meant to match the axes of choice, motion input, and color input. The axis of choice for each context was determined by averaging the top right eigenvector determined by the Jacobian at each expansion point. The motion input axis was determined by the input weight vector corresponding to the motion input weight stream. Similarly, the color input axis was determined by the input weight vector corresponding to the color input stream. These three vectors were orthogonalized to create the subspace.  We then projected the JSLDS states and expansion points (or RNN states and numerical fixed points for the standard trained RNN) into this subspace to create the plots.

\subsubsection{Contextual integration experiment performed without JSLDS regularization.}

\begin{figure}[hbt!]
  \centering
  \includegraphics[width=\linewidth]{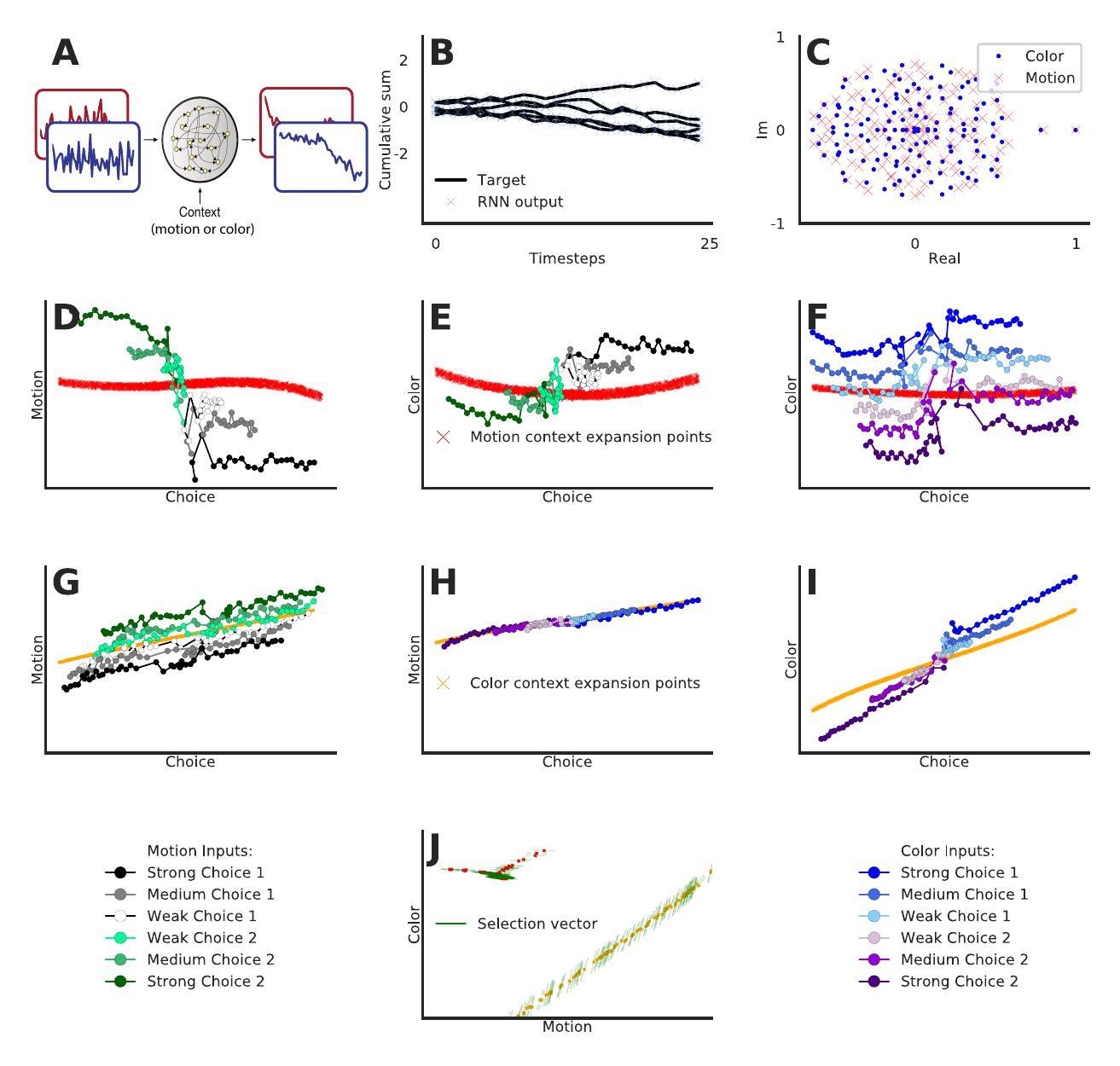}
  \vspace{-1em}
  \caption{Context-dependent integration for standard vanilla RNN (no JSLDS co-training) \textbf{A.} One of two white-noise input streams (motion or color) is  selected to be integrated based on a static context input. The other stream is ignored. \textbf{B.} Sample held-out trial outputs and targets. \textbf{C.} Typical eigenvalues at a sample fixed point (found numerically) for motion (red x's) and color (blue dots) contexts.  \textbf{D-J.} The RNN states (averaged) and fixed points are projected into the subspace spanned by the axes of choice, motion, and color. Movement along the choice axis represents integration of evidence and the relevant input stream deflects along the relevant input axis. The input axes of \textbf{E,F,G} have been intensified. The trials used in \textbf{F} and \textbf{G} are the same trials as \textbf{D-E} and \textbf{H-I}, respectively, but re-sorted and averaged according to the direction and strength of the irrelevant input. The fixed points were computed separately for motion (red x's) and color contexts (orange x's). \textbf{J.} Global arrangement of the selection vectors (green lines) and line attractor fixed points for both contexts projected onto the input axes. Inputs are selected by the selection vector (which is approximately orthogonal to the contextually irrelevant input) and integrated along the line attractor.}
  \label{context_int_nojslds_fig}
\end{figure}

We repeated the contextual integration experiment with a standard trained vanilla RNN without the JSLDS co-training. After training, we numerically found its fixed points for both contexts and recreated the plot from the main paper in Figure \ref{context_int_nojslds_fig}.  We see that the standard trained RNN finds basically the same solution as the co-trained networks. The standard trained network eigenvalues tend to exhibit larger imaginary components, but the top eigenvalue for both contexts is still $(1,0)$. So for the contextual integration experiment with a vanilla RNN the JSLDS does not seem to dramatically change the fixed point solution, although as observed in the main paper it still significantly improves the linearized approximation of the dynamics. 

\clearpage

\subsection{Monkey reach task}
\subsubsection{Experimental details}
\label{monkey_methods}

\begin{figure}[h]
  \centering
  \includegraphics[width=1.0\linewidth]{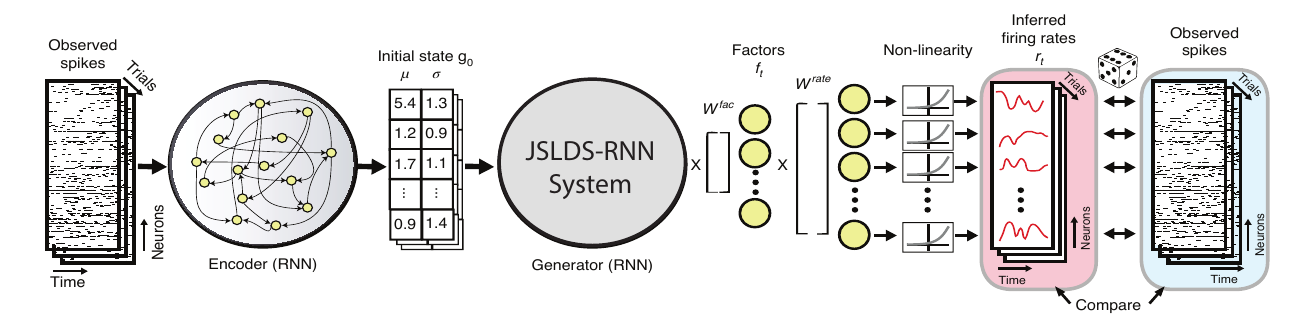}
  \caption{The LFADS-JSLDS architecture. The JSLDS-RNN system is used as the generator. After training, the model can produce firing rates from either the JSLDS or RNN generator. }
  \label{lfads_jslds_fig}
\end{figure}

The data consists of 2296 trials of spiking activity recorded from 202 neurons simultaneously while a monkey made reaching movements during a maze task across 108 reaching conditions. The analyzed trials were 900-ms long and to train the model we used a bin size of 5 ms.

For the LFADS-JSLDS model (Fig. \ref{lfads_jslds_fig}),  we used a 100 unit GRU for the RNN in the generator. To train LFADS-JSLDS, one simply includes the JSLDS loss function in the LFADS loss. We used the Adam optimizer with default settings. See Table \ref{monkeytable} for important hyperparameter settings.  

\begin{table}[h]
  \caption{Hyperparameters used in monkey reach task}
  \label{monkeytable}
  \centering
  \begin{tabular}{lll}
    \toprule
    \cmidrule(r){1-2}
    Model     & LFADS-JSLDS & LFADS \\
    \midrule
    RNN type & GRU  & GRU \\
    Generator Dimension           & 100 & 100 \\
    Encoder Dimension             & 100 & 100 \\
    Factors Dimension             & 40  & 40 \\
    Keep probability              & .98 & .98 \\
    Bin size                      & 5   & 5 \\
    Batch size                    & 128 & 128 \\
    Initial learning rate         & .05 & .05 \\
    L2 regularization             &2.0e-2 & 2.0e-2 \\
    Expansion network layers      & 2   & n/a \\
    Expansion network units/layer & 100 & n/a \\
    Expansion network activation  & tanh & n/a \\
    $\lambda_{\mathsf{RNN}}$      & 1.0 & n/a   \\
    $\lambda_{\mathsf{JSLDS}}$    & 1.0  & n/a \\
    $\lambda_{e}$                 & 100.0 &n/a \\
    $\lambda_{a}$                 & 20.0 & n/a\\
    \bottomrule
  \end{tabular}
\end{table}

\subsubsection{Subspace Analysis}
\label{monkey_methods_subspace}
We perform a subspace analysis to compare the JSLDS analysis to a jPCA analysis. The jPCA method finds linear combinations of principal components that capture rotational structure in data. Through a series of steps, it finds a transformation between a neural system at each timestep and its temporal derivative. The subspace angle refers to the angle between the planes defined by the top principal components from the jPCA analysis and the planes defined by the top JSLDS eigenvectors. To be concrete, associated with each conjugate pair of complex eigenvalues from the jPCA analysis is a conjugate pair of principal components that define a plane. Analogously, for each of the top complex pairs of eigenvalues from the JSLDS dynamics matrix, a corresponding conjugate pair of complex eigenvectors also define a plane. The subspace angle measures how similar these planes are and can be used to match up the corresponding jPCA eigenvalues (constrained to be complex) and the JSLDS eigenvalues. This is the connection displayed in Figure 4E. The fact that the top eigenvalues of the JSLDS match up with the corresponding jPCA eigenvalues indicates our method is working correctly.

\subsubsection{LFADS experiment performed without JSLDS}
\label{monkey_methods_no_jslds}

\begin{figure}[h]
  \centering
  \includegraphics[width=1.0\linewidth]{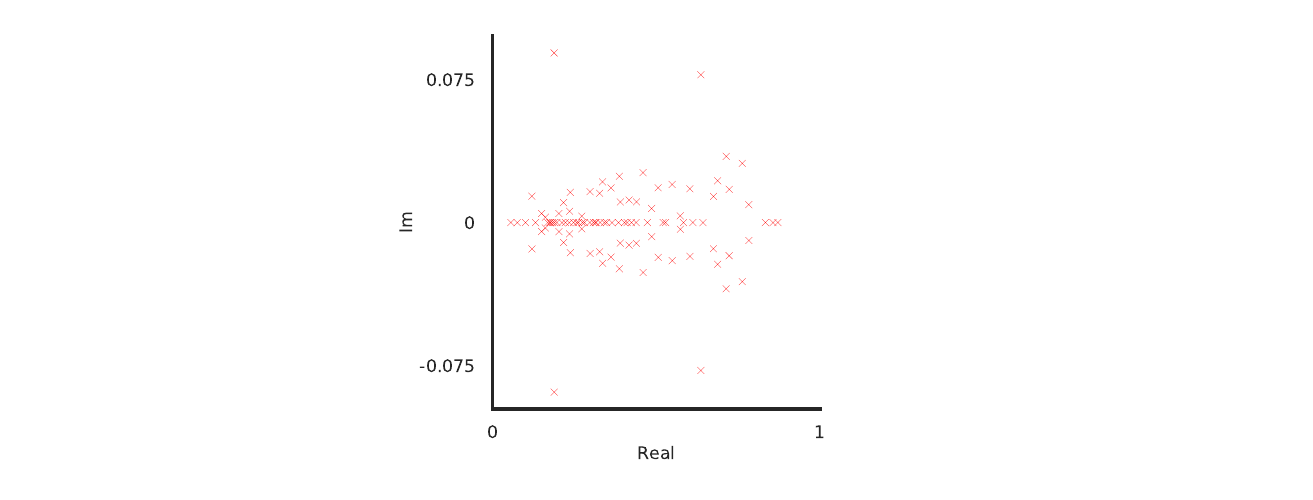}
  \caption{Eigenvalues of the trained LFADS (without JSLDS co-training) RNN generator's Jacobian at its single fixed point.}
  \label{lfads_no_jslds_evals}
\end{figure}

We also compared the fixed point solution for the LFADS model trained without the JSLDS co-training. We trained the LFADS with the exact same hyperparameters as the LFADS-JSLDS except without the JSLDS co-training related terms. We observed this setup also learned a single linear system (Fig. \ref{lfads_no_jslds_evals}). So it seems in this case the JSLDS co-training did not have a significant effect on the fixed point solution.

%%%%%%%%%%%%%%%%%%%%%%%%%%%%%%%%%%%%%%%%%%%%%%%%%%%%%%%%%%%%
\subsection{NeurIPS Checklist}
\begin{enumerate}

\item For all authors...
\begin{enumerate}
  \item Do the main claims made in the abstract and introduction accurately reflect the paper's contributions and scope?
    \answerYes{}{}
  \item Did you describe the limitations of your work?
    \answerYes{}
  \item Did you discuss any potential negative societal impacts of your work?
    \answerNA{}{See broader impact statement.}
  \item Have you read the ethics review guidelines and ensured that your paper conforms to them?
    \answerYes{}{}
\end{enumerate}

\item If you are including theoretical results...
\begin{enumerate}
  \item Did you state the full set of assumptions of all theoretical results?
    \answerNA{}{}
	\item Did you include complete proofs of all theoretical results?
    \answerNA{}{}
\end{enumerate}

\item If you ran experiments...
\begin{enumerate}
  \item Did you include the code, data, and instructions needed to reproduce the main experimental results (either in the supplemental material or as a URL)?
    \answerYes{We have linked to a github repository containg the code and two Google Colab notebooks that walk through the model for the synthetic tasks. We are unable to provide the monkey data as it is proprietary.}
  \item Did you specify all the training details (e.g., data splits, hyperparameters, how they were chosen)?
    \answerYes{}{See Appendix.}
	\item Did you report error bars (e.g., with respect to the random seed after running experiments multiple times)?
    \answerYes{}{}
	\item Did you include the total amount of compute and the type of resources used (e.g., type of GPUs, internal cluster, or cloud provider)?
    \answerYes{}{This information is available in the included Colab notebooks}
\end{enumerate}

\item If you are using existing assets (e.g., code, data, models) or curating/releasing new assets...
\begin{enumerate}
  \item If your work uses existing assets, did you cite the creators?
    \answerYes{}{}
  \item Did you mention the license of the assets?
    \answerNA{}{}
  \item Did you include any new assets either in the supplemental material or as a URL?
    \answerYes{}{}{}
  \item Did you discuss whether and how consent was obtained from people whose data you're using/curating?
    \answerNA{}{}
  \item Did you discuss whether the data you are using/curating contains personally identifiable information or offensive content?
    \answerNA{}{}
\end{enumerate}

\item If you used crowdsourcing or conducted research with human subjects...
\begin{enumerate}
  \item Did you include the full text of instructions given to participants and screenshots, if applicable?
    \answerNA{}{}
  \item Did you describe any potential participant risks, with links to Institutional Review Board (IRB) approvals, if applicable?
    \answerNA{}{}
  \item Did you include the estimated hourly wage paid to participants and the total amount spent on participant compensation?
    \answerNA{}{}
\end{enumerate}

\end{enumerate}

%%%%%%%%%%%%%%%%%%%%%%%%%%%%%%%%%%%%%%%%%%%%%%%%%%%%%%%%%%%%

\end{document}